\documentclass[letterpaper]{article} 
\usepackage{aaai23}  
\usepackage{times}  
\usepackage{helvet}  
\usepackage{courier}  
\usepackage[hyphens]{url}  
\usepackage{graphicx} 
\usepackage{natbib}  
\usepackage{caption} 
\usepackage{stfloats}

\nocopyright
\usepackage{algorithm}
\usepackage{algorithmic}
\usepackage{caption}
\usepackage{subcaption}
\usepackage{multirow}
\usepackage{graphicx}
\usepackage{array}
\usepackage{mathtools}
\usepackage{wrapfig}
\usepackage{amsmath}
\usepackage{newfloat}
\usepackage{listings}
\DeclareCaptionStyle{ruled}{labelfont=normalfont,labelsep=colon,strut=off} 
\floatstyle{ruled}
\newfloat{listing}{tb}{lst}{}
\floatname{listing}{Listing}
\usepackage{color}

\title{Can LLMs be Good Financial Advisors?: An Initial Study in Personal Decision Making for Optimized Outcomes}
\author{
Kausik Lakkaraju, 
Sai Krishna Revanth Vuruma, 
Vishal Pallagani, \\
Bharath Muppasani,  
Biplav Srivastava
}
\affiliations{
    University of South Carolina
}

\usepackage{bibentry}

\begin{document}

\maketitle

\begin{abstract}
Increasingly powerful Large Language Model (LLM) based chatbots, like ChatGPT and Bard, are becoming available to users that have the potential to revolutionize the quality of decision-making achieved by the public. In this context, we set out to investigate how such systems perform in the personal finance domain, where financial inclusion has been an overarching stated aim of banks for decades. 
We asked 13 questions representing banking products in personal finance: bank account, credit card and certificate of deposits and their inter-product interactions, and decisions related to  high-value purchases, payment of bank dues, and investment advice, and in different dialects and languages (English, African American Vernacular English, and Telugu). We find that although the outputs of the chatbots are fluent and plausible, there are still critical gaps in providing accurate and reliable financial information using LLM-based chatbots.
\end{abstract}

\section{Introduction}

\begin{table*}[hbt!]
{\tiny
\centering
\begin{tabular}{|m{6em}|m{2em}|m{24em}|m{24em}|m{17.5em}|}
    \hline
    {Product Interactions} &
    {Query Identifier} &
    {Queries} &
    {Variables with their values} &
    {Constraints}
    \\ \hline

    \multirow{4}{6em}{CC} &
    {Q1} &
    {I am making a \textbf{purchase of \$1000} using my credit card. My \textbf{billing cycle is from March 25th to April 24th}. Today is March 31st, and I have a \textbf{due of \$2000} on my account. My total \textbf{credit line is \$2,800}. Would you recommend I make the purchase now or later in the future?} &
    {$x_{PA}$ = 1000, $x_{BC}$ = (March 25th - April 24th), $x_{DA}$ = 2000, $x_{CL}$ = 2800} &
    \multirow{7}{17.5em}
    {\begin{equation}
    x_{DA} + x_{PA} < x_{CL}
    \end{equation}} \\ \cline{2-4}

    &
    {Q2} &
    {I am making a \textbf{purchase of \$1000} using my credit card. My \textbf{billing cycle is from March 25th to April 24th}. Today is March 31st, and I have a \textbf{due of \$2000} on my account. My total \textbf{credit line is \$3,800}. Would you recommend I make the purchase now or later in the future?} &
    {$x_{PA}$ = 1000, $x_{BC}$ = (March 25th - April 24th), $x_{DA}$ = 2000, $x_{CL}$ = 3800} &
    \\ \cline{2-4}

    &
    {Q3} &
    {I get \textbf{5\% cashback} if I buy furniture using my credit card. I am \textbf{buying furniture worth \$1000} using my credit card. My \textbf{billing cycle is from March 25th to April 24th}. Today is March 31st, and I have a \textbf{due of \$2000} on my account. My total \textbf{credit line is \$2,800}. Would you recommend I make the purchase now or later in the future?} &
    {$x_{CP}$ = 5\%, $x_{PA}$ = 1000, $x_{BC}$ = (March 25th - April 24th), $x_{DA}$ = 2000, $x_{CL}$ = 2800} &
     \\ \cline{2-4}

    &
    {Q4} &
    {I get \textbf{5\% cashback} if I buy furniture using my credit card. I am \textbf{buying furniture worth \$1000} using my credit card. My \textbf{billing cycle is from March 25th to April 24th}. Today is March 31st, and I have a \textbf{due of \$2000} on my account. My total \textbf{credit line is \$3,800}. Would you recommend I make the purchase now or later in the future?} &
     {$x_{CP}$ = 5\%, $x_{PA}$ = 1000, $x_{BC}$ = (March 25th - April 24th), $x_{DA}$ = 2000, $x_{CL}$ = 3800} &
    \\ \cline{1-4}

    {CC (AAVE)} &
    {Q5} &
    {I be makin' a \textbf{purchase of \$1000} usin' i's credit card. I's \textbf{billin' cycle be from march 25th to april 24th}. Today be march 31ts, and i done a \textbf{due of \$2000} on i's account. I's total \textbf{credit line be \$2,800}. Would you recommend i make de purchase now o lateh in de future?} &
    {$x_{PA}$ = 1000, $x_{BC}$ = (March 25th - April 24th), $x_{DA}$ = 2000, $x_{CL}$ = 2800} &
     \\ \cline{1-4}

    \multirow{2}{6em}{CC (Telugu)} &
    {Q6} &
    \begin{minipage}{.15\textwidth}
          \vspace{0.8mm}
          \centering
          \includegraphics[width=2\textwidth,bb=0 0 300 80]{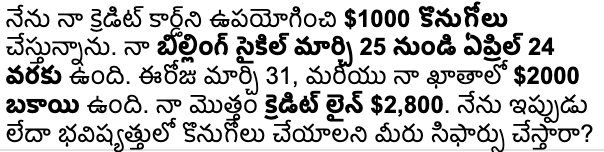}
    \end{minipage} &
     {$x_{PA}$ = 1000, $x_{BC}$ = (March 25th - April 24th), $x_{DA}$ = 2000, $x_{CL}$ = 2800} & 
     \\ \cline{2-4}

    & 
    {Q7} &
    \begin{minipage}{.15\textwidth}
          \vspace{0.6mm}
          \centering
        \includegraphics[width=2\textwidth,bb=0 0 300 80]{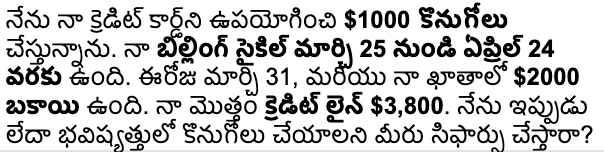}
    \end{minipage} &
    {$x_{PA}$ = 1000, $x_{BC}$ = (March 25th - April 24th), $x_{DA}$ = 2000, $x_{CL}$ = 3800}&
    \\ \hline
    
    \multirow{2}{6em}{CC and AB} &
    {Q8} &
    {I am making a \textbf{purchase of \$1000} using my credit card. My \textbf{billing cycle is from March 25th to April 24th}. Today is March 31st, and I have a \textbf{due of \$2000} on my account. My total \textbf{credit line is \$3,800}. I have \textbf{\$10,000 in my bank} which I can use to pay my credit card balance any time. Would you recommend I make the purchase now or later in the future?} &
    {$x_{PA}$ = 1000, $x_{BC}$ = (March 25th - April 24th), $x_{DA}$ = 2000, $x_{CL}$ = 3800, $x_{AB}$ = 10000} &
    \multirow{2}{17.5em}
    {Constraint (1) must be satisfied. In addition, if the user chooses to pay the due immediately, the following constraints must also hold true. 
    \begin{equation}
    x_{DA}  < x_{AB}
    \end{equation}

    \begin{equation}
    ,x_{PA}  < x_{CL}
    \end{equation}}
    \\ \cline{2-4}

    &
    {Q9} &
    {I get \textbf{5\% cashback} if I buy furniture using my credit card. I am \textbf{buying furniture worth \$1000} using my credit card. My \textbf{billing cycle is from March 25th to April 24th}. Today is March 31st, and I have a \textbf{due of \$2000} on my account. My total \textbf{credit line is \$3,800}. I have \textbf{\$10,000 in my bank} which I can use to pay my credit card balance any time. Would you recommend I make the purchase now or later in the future?} &
    {$x_{CP}$ = 5\%, $x_{PA}$ = 1000, $x_{BC}$ = (March 25th - April 24th), $x_{DA}$ = 2000, $x_{CL}$ = 3800, $x_{AB}$ = 10000} &
    \\ \hline

    \multirow{2}{6em}{CC and CD} &
    {Q10} &
    {I have a credit card \textbf{due of \$2800}. The total \textbf{credit line is \$2800}. If I don't pay a \textbf{minimum of \$100} by the end of billing cycle, my \textbf{APR would be 27\%}. If I pay the minimum amount by the end of billing cycle, \textbf{APR will be 25\%}. My \textbf{billing cycle is from March 25th to April 24th}. Today is March 31st. If I choose to deposit some amount as certificate of deposit (CD), I will get an \textbf{interest of 6\% on the amount deposited}. Do you recommend I pay the full credit card due or do a certificate of deposit or pay my due and deposit the rest?} &
    {$x_{APR}$ = 27\% (with late fee) and 25\% without late fee, $x_{MD}$ = 100, $x_{BC}$ = (March 25th - April 24th), $x_{DA}$ = 2800, $x_{CL}$ = 2800, $x_{CDP}$ = 6\%} &
    \multirow{2}{17.5em}
    {\begin{equation}
    x_{DA}  < x_{CL}
    \end{equation}
    AB was not provided in this query. So we cannot specify any additional constraints in this case from the given data.
    }\\ \cline{2-4}

    &
    {Q11} &
    {I have a credit card \textbf{due of \$2800}. The total \textbf{credit line is \$3800}. If I don't pay a \textbf{minimum of \$100} by the end of billing cycle, my \textbf{APR would be 27\%}. If I pay the minimum amount by the end of billing cycle, \textbf{APR will be 25\%}. My \textbf{billing cycle is from March 25th to April 24th}. Today is March 31st. If I choose to deposit some amount as certificate of deposit (CD), I will get an \textbf{interest of 6\% on the amount deposited}. Do you recommend I pay the full credit card due or do a certificate of deposit or pay my due and deposit the rest?} &
    {$x_{APR}$ = 27\% (with late fee) and 25\% without late fee, $x_{MD}$ = 100, $x_{BC}$ = (March 25th - April 24th), $x_{DA}$ = 2800, $x_{CL}$ = 3800, $x_{CDP}$ = 6\%} &
     \\ \hline

    \multirow{2}{6em}{CC, CD and AB} &
    {Q12} &
    {I have a credit card \textbf{due of \$2800}. The total \textbf{credit line is \$2800}. If I don't pay a \textbf{minimum of \$100} by the end of billing cycle, my \textbf{APR would be 27\%}. If I pay the minimum amount by the end of billing cycle, \textbf{APR will be 25\%}. My \textbf{billing cycle is from March 25th to April 24th}. Today is March 31st. I currently have \textbf{\$2,800 in my personal checking account}. If I choose to deposit some amount as certificate of deposit (CD), I will get an interest of \textbf{6\% on the amount deposited}. Do you recommend I pay the full credit card due with my personal account balance or do a certificate of deposit or pay my due and deposit the rest?} &
    {$x_{APR}$ = 27\% (with late fee) and 25\% without late fee, $x_{MD}$ = 100, $x_{BC}$ = (March 25th - April 24th), $x_{DA}$ = 2800, $x_{CL}$ = 2800, $x_{CDP}$ = 6\%, $x_{AB}$ = 2800} &
    \multirow{2}{17.5em}{
    \begin{equation}
    \begin{aligned}
         [(x_{DA} - x_{MD}) * x_{APR} \leq \\
          (x_{AB} - x_{MD}) * x_{CDP}]
    \end{aligned}
    \end{equation}

    \begin{equation}
    ,[(x_{AB} - x_{DA}) > 0]
    \end{equation} 
    } \\ \cline{2-4}
 
    &
    {Q13} &
    {I have a credit card \textbf{due of \$2800}. The total \textbf{credit line is \$2800}. If I don't pay a \textbf{minimum of \$100} by the end of billing cycle, my \textbf{APR would be 27\%}. If I pay the minimum amount by the end of billing cycle, \textbf{APR will be 25\%}. My \textbf{billing cycle is from March 25th to April 24th}. Today is March 31st. I currently have \textbf{\$3,800 in my personal checking account}. If I choose to deposit some amount as certificate of deposit (CD), I will get an \textbf{interest of 6\% on the amount deposited}. Do you recommend I pay the full credit card due with my personal account balance or do a certificate deposit or pay my due and deposit the rest?} &
    {$x_{APR}$ = 27\% (with late fee) and 25\% without late fee, $x_{MD}$ = 100, $x_{BC}$ = (March 25th - April 24th), $x_{DA}$ = 2800, $x_{CL}$ = 2800, $x_{CDP}$ = 6\%, $x_{AB}$ = 3800}&
     \\ \hline
\end{tabular}
\caption{Table showing different product interaction categories considered, query identifiers, the queries posed to the chatbots under each category, variables used in each query with their corresponding chosen values and constraints the chatbots need to consider while answering the user queries.}
\label{tab:queries}
}
\end{table*}

Consider a freshman that has just started making personal financial decisions. They open a bank account to save up money and get their first credit card. They are given some seed money by their family and they also start earning by working on campus.
The student is encouraged by their support system to start thinking about saving into products like Certificate of Deposits (CDs) that earn higher interest. As the student makes a series of decisions in their academic and subsequent professional life, they need to make sound financial decisions and may look for resources online to assist them. An optimal decision needs to consider how the banking products interact with each other along with the changing needs of the student.

For users like this student, increasingly powerful LLM-based chatbots that have the potential to revolutionize the quality of
decision for personal finance are becoming available. LLMs have demonstrated tremendous potential across diverse domains \cite{zhao2023survey}, such as natural language processing \cite{min2021recent} and protein structure \cite{hu2022protein}, and have been claimed to show sparks of artificial general intelligence \cite{bubeck2023sparks}. These models have been implemented in several applications, ranging from mental health assistants \cite{xiao2023powering} to financial advisement \cite{yue2023gptquant}. In the finance domain, LLMs have been used to develop applications such as fraud detection, risk management, and financial forecasting \cite{pfpt-paper}. They have been used to analyze financial data, predict stock prices, and generate automated reports. However, with the advent of recent models such as OpenAI's ChatGPT, Google's Bard, and BloombergGPT \cite{wu2023bloomberggpt}, a comparative chatbot study is needed to evaluate their ability to be financial advisors. In this paper, we present an initial study of ChatGPT and Bard in providing personal decision-making for optimized outcomes. 

It is widely known that LLMs based systems have unique limitations. 
For example, they may struggle with common-sense reasoning tasks \cite{li2022systematic}, encounter challenges when handling symbols \cite{frieder2023mathematical}, and are susceptible to hallucinations \cite{bang2023multitask}. 



With this work, we make the following contributions:
\begin{itemize}
\item identify a personal financial planning scenario involving a series of tasks (plans) and optimization of decisions.
\item show how leading LLM-based chatbots perform in them and analyze their behavior.
\item lay out challenges that future chatbots in this area should overcome to provide trusted financial recommendations. 
\end{itemize}





We thus highlight  the potential and  limitations of current LLM-based systems - ChatGPT and Bard - in their role as financial advisors. We included all the queries posed and responses from both ChatGPT and Bard in our GitHub repository\footnote{https://github.com/ai4society/LLM-CaseStudies/tree/main/Finance} along with a few snapshots of the actual conversations.

\clearpage
\section{Personal Finance Use Case}

\subsection{Setup: Tools and Procedure}
\subsubsection{Chatbots Tested}
\begin{enumerate}
    \item \textbf{ChatGPT:} ChatGPT \cite{openai2023gpt4} is an LLM-based chatbot created by OpenAI that was trained on large amount of text data from the internet, including books and articles. ChatGPT is capable of answering questions, generating text and converse with users in a natural way. It can also learn from users and adapt to new information.
    \item \textbf{Bard:} Bard \cite{google-bard} is an LLM-based chatbot created by Google that was trained on large amount of text data and is capable of generating human-like text in response to user prompts and queries. Like ChatGPT, it is also capable of conversing with users about wide variety of topics in a natural way and adapt to new information. 
\end{enumerate}

\subsubsection{Product Interaction Categories}
Product interaction refers to interaction between different products like Credit Card (CC), Certificate of Deposit (CD) and Account Balance (AB). 
Each product has different quantitative properties. For example, credit card due, limit line and billing cycle are some of the properties that would provide credit card information (not private information) of the user. Different properties pertaining to these products are:
\begin{itemize}
    \item \textbf{Purchase Amount (PA)}: It is the amount spent by the user on purchase of a product. 
    \item \textbf{Billing Cycle (BC)}: It is the billing cycle of user's credit card. 
    \item \textbf{Due Amount (DA)}: The amount that is due on the user's credit card for the specified billing cycle. 
    \item \textbf{Credit Line (CL)}: The maximum amount that user could spend using their credit card. If the amount spent exceeds this value, the credit card company could charge additional interest.
    \item \textbf{Cashback Percentage (CP)}: The \% of amount which will be returned to the user in the form of cashback on buying furniture using their credit card. 
    \item \textbf{Account Balance (AB)}: The amount of cash present in user's personal bank account.
    \item \textbf{Annual Percentage Rate (APR)}: The APR is charged if there is due on the credit card after the due date. Some financial institutions choose to charge a late fee if the \textbf{minimum due (MD)} is not paid. It is calculated by the formula, Daily Period Rate (DPR) x Billing Cycle (in days) x Average Daily Balance (ADB).
    \item \textbf{Certificate of Deposit Percentage (CDP)}: The \% of interest accumulated on the cash deposited by the user in the form of CD.
\end{itemize}
Based on different combinations of these products, we classified the queries into 4 categories. These four categories along with the queries posed under each category, the variables used in each query and the constraints the chatbot has to take into consideration to make a sound recommendation are shown in Table \ref{tab:queries}. In the CC category, we considered a different dialect of English called African American Vernacular English (AAVE) and Telugu, one of the well-known languages from India, to observe how the chatbots handle queries in a different language or dialect.

\subsection{Findings}
In this subsection, we present the findings from the interesting (and sometimes insightful) conversations we had with Bard and ChatGPT.
\subsubsection{Differences Between the Chatbots}
Table \ref{tab:diff-llms} shows the differences that were identified between Bard and ChatGPT when queries listed out in Table \ref{tab:queries} were asked. We compare these models on various criteria related to their performance in answering queries. The criteria include accuracy, utilization of user information, personalized suggestions, use of visual aids, bias in recommendations, provision of multiple response drafts, learning from mistakes, and understanding of different dialects and languages.

\begin{figure*}[h]
 \centering
   \includegraphics[scale=0.65]{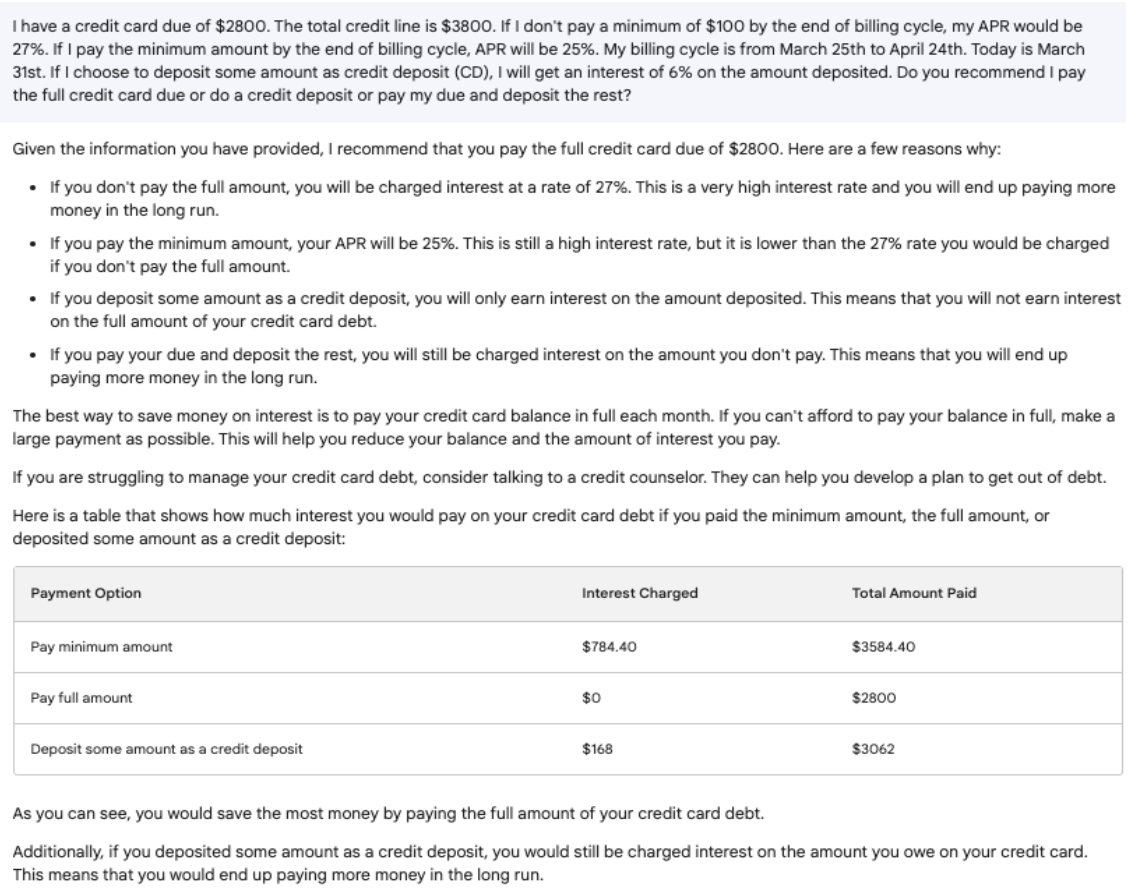}  
   \caption{Bard Response (Q11-AB11).}
  \label{fig:sample-bard}
\end{figure*}

\begin{figure*}[!tbp]
 \centering
   \includegraphics[scale=0.45]{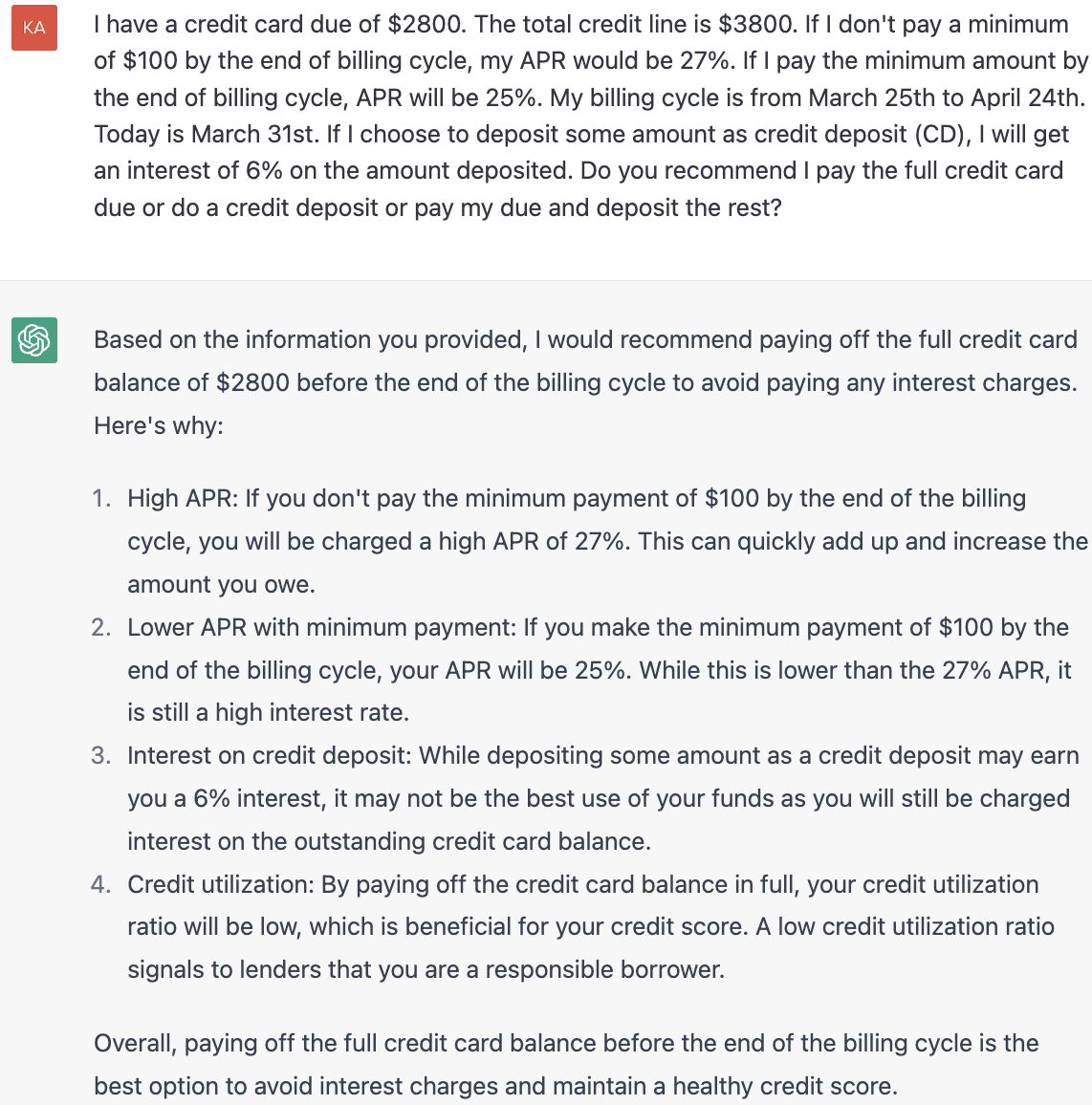}  
   \caption{ChatGPT Response (Q11-AC11).}
  \label{fig:sample-gpt}
\end{figure*}

\begin{table}
{\tiny
\centering
 \begin{tabular}{|m{1.2em}|m{15.1em}|m{15.1em}|}
    \hline
          {S.No.} &    
          {Bard} & 
          {ChatGPT} \\ \hline

          {1.} &    
          {Bard gives accurate results if the question is asked directly (for ex., \$2,250 x 0.0006849 x 30 = \$46.23075.)} & 
          {ChatGPT gives inaccurate results if the question is asked directly (\$2,250 x 0.0006849 x 30 = \$46.90 (rounded to the nearest cent)) } \\ \hline

          {2.} &    
          {Bard does not utilize the information the user provides completely and calculates CUR less often than ChatGPT.} & 
          {ChatGPT calculates CUR and reasons using the computed CUR more often than Bard} \\ \hline

          {3.} &    
          {Bard usually does not give personalized suggestions (especially, when the (Due + purchase amount) $>$ Credit line).} & 
          {ChatGPT gives personalized suggestions more often than Bard.} \\ \hline

          {4.} &    
          {As a response to one of the queries, Bard gave a recommendation by making use of a table with different options that user could choose from as shown in  Figure \ref{fig:sample-bard}.} & 
          {ChatGPT did not use any kind of visual aids.} \\ \hline
          
          {4.} &    
          {Bard gave biased recommendation i.e., biased towards recommending the user to make the purchase immediately (in one case, it gave only pros for buying the furniture immediately even though it has serious cons).} & 
          {ChatGPT never gave biased recommendations (it never encourages the user to buy the furniture immediately unless there is no risk involved).} \\ \hline

          {5.} &    
          {Bard gives 3 different drafts (with some changes in the response) for the same query.} & 
          {ChatGPT does not provide different drafts.} 
          \\ \hline
          
          {6.} &    
          {With each query posed, the content (calculations) of Bard is not improving as much as ChatGPT. It is not learning from its mistakes immediately.} & 
          {ChatGPT corrects its errors more often than Bard} \\ \hline

          {7.} &    
          {Bard understood African-American Vernacular English (AAVE) dialect and gave a reasonable response to the query} & 
          {When query was posed in AAVE dialect, ChatGPT did not understand it immediately. When we posed the same query again in the same dialect, it understood the query and gave a reasonable recommendation .} \\ \hline

          {8.} &    
          {Bard was not trained to understand Telugu language.} & 
          {Though ChatGPT can understand Telugu language and responds in Telugu if the user query is in Telugu, the response it generated was incomplete and had a lot of grammatical errors which made the response very hard to understand.} \\ \hline

    \end{tabular}
    \caption{Differences between the responses generated by Bard and ChatGPT when queries related to finance domain were posed.}
    \label{tab:diff-llms}
    }
\end{table}

\subsubsection{Error Categories}
We identified some limitations / errors in the responses generated by both the chatbots and classified them into the following categories:
\begin{itemize}
    \item \textbf{Lack of Personalized Recommendations:} When the agent makes a generalized recommendation without using all the information provided by the user, we consider this as lack of personalized recommendation.
    \item \textbf{Mathematical Errors:} We consider errors like rounding errors, calculation errors, etc. as mathematical errors.
    \item \textbf{Perceptual Errors:} When the agent misinterprets information given by the user or makes assumptions on unknown data, we consider these as perceptual errors.
    \item \textbf{Grammatical Errors:} We consider typos, grammatical errors, etc. as grammatical errors (we encountered these errors only in Telugu text generated by ChatGPT).
    \item \textbf{Lack of Visual Aids:} When the agent doesn't use visual aids like tables, graphs, etc. in its response, we consider these as lack of visual aids.
\end{itemize}
Table \ref{tab:errs} shows the percentage of queries for which the chatbots exhibited each of these errors. We also list out the individual query identifiers. Qi denotes the query identifier as previously defined (and also shown in Table \ref{tab:queries}). ABi and ACi refer to the corresponding Bard and ChatGPT responses respectively. 'i' denotes the identifier (number). Figures \ref{fig:sample-bard} and \ref{fig:sample-gpt} show the response generated by Bard and ChatGPT chatbots respectively. For this one query, Bard made use of a table (though it misinterpreted user information) and ChatGPT did not.

\begin{table}[!h]
{\small
\centering
\begin{tabular}{|m{6em}|m{7em}|m{4em}|m{4em}|}
    \hline
    {Error Category} &
    {Queries} &
    {\% of Bard Queries} &
    {\% of ChatGPT Queries} \\ \hline

    {Lack of Personalized Recommendations} &
    {Q1-AB1, Q3-AB3, Q3-AC3, Q4-AB4, Q5-AB5, Q6-AC6, Q7-AC7, Q8-AB8, Q9-AB9, Q10-AC10, Q11-AC11, Q12-AB12, Q12-AC12, Q13-AB13} &
    {53.84\%}  &
    {46.15\%}  \\ \hline
    
    {Mathematical Errors} &
    {Q2-AB2, Q9-AC9, Q10-AB10} &
    {15.38\%} &
    {7.69\%} \\ \hline

    {Perceptual Errors} &
    {Q8-AC8, Q10-AB10, Q11-AB11} &
    {15.38\%} &
    {7.69\%} \\ \hline

    {Grammatical Errors} &
    {Q6-AC6, Q7-AC7} &
    {0\%} &
    {15.38\%$^*$} \\ \hline

    {Lack of Visual Aids} &
    {All except Q11-AB11} &
    {92.30\%} &
    {100\%} \\ \hline
    
\end{tabular}
\caption{Table showing \% of queries for which the chatbots exhibited different errors along with individual query-response identifiers. `Qi' denotes the query identifier, `ABi' and `ACi' represent the corresponding Bard and ChatGPT responses respectively where `i' is the identifier.}
\label{tab:errs}
}
\end{table}



\section{Discussion and Conclusion}

The application of language models in the finance industry has witnessed a surge in recent times due to their ability to process vast volumes of unstructured data and extract valuable insights. This paper delves into the performance of two prominent language models, Bard and ChatGPT, within the finance domain.

We also find the following challenges in evaluating LLM-based systems for finance domains:
\begin{itemize}
    \item C1: Changing nature of answers for the same question. How does one create reference test cases since the answers change over time? 
    \item C2: Inability of the chatbots to do numeric reasoning 
    \item C3: Presenting results with easy to follow graphics.
    \item C4: Support for languages used by customers from different population groups. We considered AAVE - (African American Vernacular English) and Telugu, an Indian language spoken by nearly 100m people world-wide. 
    \item C5: Evaluation the response of users from a diverse set of background. We only considered college students in this study.
\end{itemize}

C1 can be mitigated by carefully cataloging  questions and system answers by identifiers that account for changing behavior over time. For C2, integration with numeric solvers like Wolfram  may help \cite{chatgpt-wolfram} although this makes the systems non-learnable over time. For C3, different data presentation strategies need to be tried. For C4, the LLM models or the chatbots need to be enhanced. For C5, more experiments are needed with inputs carefully modeling the characteristics of the different user groups. These are just preliminary  challenges and we expect them to grow as more researchers will try LLM-based systems in complex and diverse application scenarios.

While our study only comprised thirteen queries, we meticulously selected them to cover various categories of credit card finance. However, there exists ample scope for more extensive testing of these chatbots by expanding the number of queries under each category or including additional categories like student loans and stock purchases. By doing so, we can gain a better understanding of the efficacy of language models in different financial domains and improve their functionality in real-world scenarios.


\clearpage
\bibliography{references}

\begin{thebibliography}{14}
\providecommand{\natexlab}[1]{#1}

\bibitem[{Alberto~Pozanco(2022)}]{pfpt-paper}
Alberto~Pozanco, D.~B., Kassiani~Papasotiriou. 2022.
\newblock PFPT: a Personal Finance Planning Tool by means of Heuristic Search
  and Automated Planning.

\bibitem[{Bang et~al.(2023)Bang, Cahyawijaya, Lee, Dai, Su, Wilie, Lovenia, Ji,
  Yu, Chung et~al.}]{bang2023multitask}
Bang, Y.; Cahyawijaya, S.; Lee, N.; Dai, W.; Su, D.; Wilie, B.; Lovenia, H.;
  Ji, Z.; Yu, T.; Chung, W.; et~al. 2023.
\newblock A multitask, multilingual, multimodal evaluation of chatgpt on
  reasoning, hallucination, and interactivity.
\newblock \emph{arXiv preprint arXiv:2302.04023}.

\bibitem[{Bubeck et~al.(2023)Bubeck, Chandrasekaran, Eldan, Gehrke, Horvitz,
  Kamar, Lee, Lee, Li, Lundberg et~al.}]{bubeck2023sparks}
Bubeck, S.; Chandrasekaran, V.; Eldan, R.; Gehrke, J.; Horvitz, E.; Kamar, E.;
  Lee, P.; Lee, Y.~T.; Li, Y.; Lundberg, S.; et~al. 2023.
\newblock Sparks of artificial general intelligence: Early experiments with
  gpt-4.
\newblock \emph{arXiv preprint arXiv:2303.12712}.

\bibitem[{Frieder et~al.(2023)Frieder, Pinchetti, Griffiths, Salvatori,
  Lukasiewicz, Petersen, Chevalier, and Berner}]{frieder2023mathematical}
Frieder, S.; Pinchetti, L.; Griffiths, R.-R.; Salvatori, T.; Lukasiewicz, T.;
  Petersen, P.~C.; Chevalier, A.; and Berner, J. 2023.
\newblock Mathematical capabilities of chatgpt.
\newblock \emph{arXiv preprint arXiv:2301.13867}.

\bibitem[{Google(2023)}]{google-bard}
Google, . 2023.
\newblock Google BARD.
\newblock In \emph{https://bard.google.com/}.

\bibitem[{Hu et~al.(2022)Hu, Xia, Zheng, Tan, Huang, Xu, and
  Li}]{hu2022protein}
Hu, B.; Xia, J.; Zheng, J.; Tan, C.; Huang, Y.; Xu, Y.; and Li, S.~Z. 2022.
\newblock Protein Language Models and Structure Prediction: Connection and
  Progression.
\newblock \emph{arXiv preprint arXiv:2211.16742}.

\bibitem[{Li et~al.(2022)Li, Kuncoro, Hoffmann, de~Masson~d’Autume, Blunsom,
  and Nematzadeh}]{li2022systematic}
Li, X.~L.; Kuncoro, A.; Hoffmann, J.; de~Masson~d’Autume, C.; Blunsom, P.;
  and Nematzadeh, A. 2022.
\newblock A systematic investigation of commonsense knowledge in large language
  models.
\newblock In \emph{Proceedings of the 2022 Conference on Empirical Methods in
  Natural Language Processing}, 11838--11855.

\bibitem[{Min et~al.(2021)Min, Ross, Sulem, Veyseh, Nguyen, Sainz, Agirre,
  Heinz, and Roth}]{min2021recent}
Min, B.; Ross, H.; Sulem, E.; Veyseh, A. P.~B.; Nguyen, T.~H.; Sainz, O.;
  Agirre, E.; Heinz, I.; and Roth, D. 2021.
\newblock Recent advances in natural language processing via large pre-trained
  language models: A survey.
\newblock \emph{arXiv preprint arXiv:2111.01243}.

\bibitem[{OpenAI(2023)}]{openai2023gpt4}
OpenAI. 2023.
\newblock GPT-4 Technical Report.
\newblock arXiv:2303.08774.

\bibitem[{Wolfram(2023)}]{chatgpt-wolfram}
Wolfram, S. 2023.
\newblock ChatGPT Gets Its “Wolfram Superpowers”!
\newblock
  \emph{https://writings.stephenwolfram.com/2023/03/chatgpt-gets-its-wolfram-superpowers/}.

\bibitem[{Wu et~al.(2023)Wu, Irsoy, Lu, Dabravolski, Dredze, Gehrmann,
  Kambadur, Rosenberg, and Mann}]{wu2023bloomberggpt}
Wu, S.; Irsoy, O.; Lu, S.; Dabravolski, V.; Dredze, M.; Gehrmann, S.; Kambadur,
  P.; Rosenberg, D.; and Mann, G. 2023.
\newblock BloombergGPT: A Large Language Model for Finance.
\newblock arXiv:2303.17564.

\bibitem[{Xiao et~al.(2023)Xiao, Liao, Zhou, Grandison, and
  Li}]{xiao2023powering}
Xiao, Z.; Liao, Q.~V.; Zhou, M.; Grandison, T.; and Li, Y. 2023.
\newblock Powering an AI Chatbot with Expert Sourcing to Support Credible
  Health Information Access.
\newblock In \emph{Proceedings of the 28th International Conference on
  Intelligent User Interfaces}, 2--18.

\bibitem[{Yue and Au(2023)}]{yue2023gptquant}
Yue, T.; and Au, C.~C. 2023.
\newblock GPTQuant's Conversational AI: Simplifying Investment Research for
  All.
\newblock \emph{Available at SSRN 4380516}.

\bibitem[{Zhao et~al.(2023)Zhao, Zhou, Li, Tang, Wang, Hou, Min, Zhang, Zhang,
  Dong, Du, Yang, Chen, Chen, Jiang, Ren, Li, Tang, Liu, Liu, Nie, and
  Wen}]{zhao2023survey}
Zhao, W.~X.; Zhou, K.; Li, J.; Tang, T.; Wang, X.; Hou, Y.; Min, Y.; Zhang, B.;
  Zhang, J.; Dong, Z.; Du, Y.; Yang, C.; Chen, Y.; Chen, Z.; Jiang, J.; Ren,
  R.; Li, Y.; Tang, X.; Liu, Z.; Liu, P.; Nie, J.-Y.; and Wen, J.-R. 2023.
\newblock A Survey of Large Language Models.
\newblock arXiv:2303.18223.

\end{thebibliography}

\end{document}